\documentclass[11pt]{article}

\usepackage{natbib}

\newenvironment{keywords}
{\bgroup\leftskip 20pt\rightskip 20pt \small\noindent{\bf Keywords:} }%
{\par\egroup\vskip 0.25ex}

\newlength\aftertitskip     \newlength\beforetitskip
\newlength\interauthorskip  \newlength\aftermaketitskip

\usepackage[a4paper,%
            left=2cm,right=2cm,top=2cm,bottom=2cm,%
]{geometry}

\newcommand{\BlackBox}{\rule{1.5ex}{1.5ex}}  % end of proof

\newtheorem{theorem}{Theorem}
\newtheorem{lemma}[theorem]{Lemma} 
 
\newtheorem{remark}[theorem]{Remark}
\newtheorem{corollary}[theorem]{Corollary}
\newtheorem{definition}[theorem]{Definition}

\usepackage[utf8]{inputenc} % allow utf-8 input
\usepackage[T1]{fontenc}    % use 8-bit T1 fonts
\usepackage{amsfonts}       % blackboard math symbols
\usepackage{nicefrac}       % compact symbols for 1/2, etc.
\usepackage{microtype}      % microtypography
\usepackage{graphicx}
\usepackage{multirow}
\usepackage{amsmath}
\usepackage{amssymb}
\usepackage{algorithm}
\usepackage{color}

\begin{document}

{\huge \bf \begin{center} Topological regularization with \\ information filtering networks \end{center}}
%\title{Learning sparse non-linear probabilistic models of complex systems with information filtering networks}
\vskip.5cm

\begin{center}{{\large Tomaso Aste} \\ \vskip10pt
Department of Computer Science, UCL, London, UK.  \\
UCL Centre for Blockchain Technologies, UCL, London, UK.  \\
Systemic Risk Centre, London School of Economics, London UK.  
}
\end{center}

%\maketitle

\begin{abstract}
A methodology to perform topological regularization via information filtering network is introduced.
This methodology can be directly applied to covariance selection problem providing an instrument for sparse probabilistic modeling with both linear and non-linear multivariate probability distributions such as the elliptical and generalized hyperbolic families. 
It can also be directly implemented for $L_0$-norm regularized multicollinear regression. 
In this paper, I describe in detail an application to sparse modeling with multivariate  Student-t. 
A specific $L_0$-norm regularized expectation-maximization likelihood maximization procedure is proposed for this sparse Student-t case.
Examples with real data from stock prices log-returns and from artificially generated data demonstrate applicability, performances, and potentials of this methodology.
%and extended to combine $L_0$, $L_1$ and $L_2$-norm regularization. 
 \\
\end{abstract}


\section{Introduction}\label{s.In}
Regularization is an important tool in machine learning to reduce the tendency of models to overfit the dataset on which they are trained and then underperform on new data. 
%- Regularization \\
In data-driven modeling, regularization typically consists in adding a penalization term to the objective function in order to control for the complexity of the model with the aim of reducing overfitting.
The idea was originally introduced by \cite{tikhonov1943stability} and since then it has permeated the field of inverse problems and machine learning.
There are different possible regularizations depending on the form of the penalization. The original Tikhonov approach (also known as ridge regression) was introduced in the context of multicollinear  regression and consisted in penalizing the sum-of-square loss function by adding the sum of square of the regression coefficients (the $L_2$-norm) giving in this way preference to models with smaller coefficients. 
%- L1 sparse modeling with lasso \\
Other forms of penalization can of course be implemented and a particularly successful one uses of the $L_1$-norm instead of the $L_2$-norm and it was named `lasso' by \cite{tibshirani1996}. 
One of the consequences of the $L_1$-norm penalization is to force certain coefficients to be set to zero producing therefore sparse models.
Sparsity is extremely advantageous for interpretability because it reduces the number of variables involved in the model.  
%- L0 modeling: the challenges \\
When sparsity and interpretability are the objective, then one would aim to penalize the objective function directly with a $L_0$-norm that introduces a cost for the number of coefficients in the model therefore directly penalizing denser models.
An advantage of the $L_0$-norm penalization is that the non-zero coefficients are not shrank in value allowing, in some cases, the use of local optimization methods.
However,  $L_0$-norm regularization has been proven to be in general challenging.
Indeed, exact optimization under $L_0$-norm penalty is computationally intractable being non-differentiable and having a combinatorically large number of possible configurations to be explored. 

% IFN
In this paper I propose the use of information filtering networks (IFN) for $L_0$-norm topological regularization. 
IFN are a class of networks originally introduced to %introduced in \cite{asteetal2005} and \cite{tumminelloetal2005} 
extract and analyze the relevant `backbone' structure of interrelations in complex systems comprising a large number of interacting elements \citep{tumminello2005tool}.
They have been shown to provide a meaningful characterization of the structure of many systems in different domains from finance  \citep{tumminelloetal2007,aste2010correlation,pozzi2013spread,musmeci2014risk,musmeci2015relation,procacci2019forecasting}, to psychology \citep{christensen2018network,christensen2018networktoolbox} and biology \citep{song2008correlation,song2012hierarchical}.
Recently \cite{TMFG} have introduced a class of IFN, named Triangulated Maximally Filtered Graph (TMFG), that is chordal and therefore particularly suited for probabilistic inference modeling. 
%These networks are  clique forests and they can be generated in a computationally efficient way.
TMFG networks are clique tree made with tetrahedra and they can be generated in a computationally efficient way.
More recently  \citep{MFCF} the TMFG approach has been radically generalized to a vaster class of clique forests with cliques of arbitrary sizes.
The algorithm was named Maximally Filtered Clique Forest (MFCF)  and it uses a computationally efficient clique expansion algorithm that has the property of being topologically invariant ensuring that the construction preserves chordality.
%The ability to efficiently use IFN for $L_0$-norm regularization introduces a tool for sparse modeling with  meaningful structures  of direct interpretability.%into the machine learning and inverse problem contexts a novel element of direct interpretability with meaningful structures. 
%- LoGo and its success with respect to lasso \\
In \cite{LoGo16} a local-global procedure (LoGo) for probabilistic modeling was introduced  using TMFG as Markov random fields. %as  inference structure in Gaussian Random Fields%within the framework to the covariance selection problem  %this issue was successfully addressed by topologically regularizing  multivariate normal models introducing a local-global procedure named LoGo. 
It was shown that in the multivariate normal case the inverse covariance is sparse and it has non-zero elements coinciding with the TMFG network edges. 
Such non-zero coefficients of the sparse inverse covariance can be computed from local inversions on the network structure. 
%In the LoGo approach, the model is sparsified accordingly with a given sparse network structure and the coefficients of the sparse inverse covariance are retrieved from local inversions on the network structure. 
This methodology was proven to be extremely effective producing sparse models with larger likelihhod performances than lasso and with lower computational burden.
%A key element of the LoGo approach is the choice of the sparse network structure. 
%In Bayesian terms this corresponds to the choice of a good prior inference model.
%- Information Filtering Networks \\
%In the original paper, LoGo utilizes a class of chordal graphs, named TMFG \citep{TMFG}, that belongs to the broader category of information filtering networks and have clique-tree structure consisting of 4-cliques separated by 3-cliques.
%In this paper I use a more general construction, named MFCF \citep{MFCF}, that allows the construction of clique forests with cliques with different sizes. 
Despite it was introduced for different purposes, the LoGo approach is a specific instance of topological  $L_0$-norm regularization. %, with TMFG, for linear modeling with multivariate normal distributions. 

%From a general perspective, 
The general problem that I am addressing in this paper can be formulated as a likelihood optimization under a topological constraint. 
In other words, the IFN is a Bayesian prior for the inference model and the posterior probability is optimized given that prior structure. 
This general problem is independent on the kind of prior inference network structure and on the kind of probability modeling.
In this paper I will show that this problem can be solved for both multivariate normal and multivariate Student-t modeling given a clique-tree prior inference network structure generated with the MFCF method. 
I will also argue that this $L_0$-norm regularization approach is applicable to the covariance selection problem and it can be therefore used for any modeling with multivariate elliptical distributions and multilinear regression problems.  
%In this paper I propose a methodology to make use of the meaningful and interpretable structure of IFN for $L_0$-norm regularization. 
%For this purpose I adopt an approach analogous to LoGo but applied to elliptical distributions and using a more general IFN construction, named MFCF \citep{MFCF}, that produces clique forests with cliques of different sizes.  
%In particular, I provide a detailed application to multivariate Student-t sparse models that are of  interest in the financial domain where probability distribution of price log-returns are often modeled with such distributions.    
To the best of my knowledge the proof of topologically-constrained likelihood optimization for Student-t models is an original result that I obtain in this paper by extending the  expectation-maximization (EM) procedure \citep{dempster1977maximum,bishop2006pattern} to this $L_0$-norm regularized sparse model.

To demonstrate applicability, robustness  and validity of this topological regularization methodology I perform a set of experiments using both synthetic and real data from financial equity prices.
I generated sets of clique tree IFNs via the MFCF approach varying the maximum  clique sizes to explore a range of different sparsities. 
The results show that there are optimal levels of sparsification for off-sample likelihood maximization.

The paper is organize as follows.
In section \ref{IFNlearning} I describe the construction of information filtering networks. 
The topological regularization approach is presented in section  \label{s.TR}.
Examples of the application of this methodology to real and synthetic data are provided in Section \ref{s.Ex}. 
Conclusions and perspectives are provided in Section \ref{s.C}.
Proof of theorems and methodological details are given in the appendixes.

%In section \ref{s.ST}, I recall the Student-t multivariate and some of its relevant properties; I also introduce some notation. 
%The construction of information filtering networks is presented in Section \ref{IFNlearning}. 
%The topological regularization methodology is  introduced in Section \ref{s.TR}.
%The construction of the $L_0$-norm IFN-penalized set of coefficients for  Student-t multivariate modeling is introduced in Section \ref{Jlearning}.
%Section \ref{Jmax} provides a demonstration that the expectation-maximization procedure can be extended to maximize likelihood of sparse Student-t models.
%Examples of the application of this methodology to real and synthetic data are provided in Section \ref{s.Ex}. 
%Conclusions and perspectives are provided in Section \ref{s.C}.
%Appendix \ref{a.St} contains some proofs.

\section{Information filtering network learning}\label{IFNlearning}

The structure of chordal  IFN  can be learned by using a clique expansion procedure as described in \cite{MFCF}, where a clique forest is constructed starting from a seed structure and including vertices into the forest one by one accordingly with a given gain function. 
The resulting network is named MFCF. 
Such a clique forest network is made of a set of cliques $\mathcal{C}$ that are the `vertices' in the clique-forest structure, the `edges' of the clique-forest structure are instead a set $\mathcal{S}$ of separators that are cliques themselves with the property that by removing one of them the connected component becomes separated into two components.
%connected into a clique forest structure through separators which belong to a separator set $ \mathcal{S}$.
Clique forests are chordal graphs.

The  MFCF network complexity can be constrained by limiting the minimum and maximum clique sizes. By increasing the clique sizes ones increases the number of edges in the network making it denser.
The full network is retrieved when the minimum clique size equals the total number of vertices.
Separators can be constrained to be unique between two cliques (multiplicity one) or to be utilizable more than once by more than two cliques (multiplicity larger than one).
The simplest clique is the 2-clique that has two elements and it is an edge. 
MFCF networks with two cliques only are segments if separators have multiplicity one or they are  maximum spanning threes when separators have arbitrary multiplicity.  
% If one limits the maximum and minimum clique sizes to two elemts only (an edge) the MFCF structure becomes the maximum spanning tree %(when separators - vertices- can be used more than once),  
The TMFG is obtained when cliques have all size 4 (tetrahedra) and the separators can be used only once.
The networks that I use in this paper have minimum clique size equal to 2 and a maximum clique sizes ranging between 2 to the total number of vertices. 

The MFCF clique expansion algorithm requires a gain function that is used to decide the inclusion of a vertex into the clique tree in a recursive way. The choice of a convenient gain function is strictly related to the problem under investigation. %There are a large number of gain function 
The gain function that I use in this paper is the sum of the squares of the coefficients of the Kendall correlation matrix. 
This gain function is a good proxy for likelihood in a range of problems. 
This is a very simple gain function that lead to networks with all cliques with maximum size.
Indeed, with this kind of additive gain the algorithm always gains by enlarging the clique, if allowed. 
I choose  Kendall correlations because they describe dependency for a broader class of multivariate random variables than the Pearson's correlations. They, are non-linear  and  have been proven to be effective in practical applications \citep{pozzi2013spread}.

The IFN structure is learned before the  maximum likelihood estimate of the model-parameters and it is passed to the optimization procedure as a Bayesian prior.
This approach is analogous to the LoGo methodology introduced in \cite{LoGo16}.
However, here we apply it to non-normal models and this has important implications. 
Indeed, outside normal modeling the structure of the IFN graph does no longer represents conditional independence and the sparse probability distribution function no longer factorizes over the IFN clique and separator structure (see \cite{lauritzen1996} and Eq.\ref{MultiVarNormalDecomp} in Appendix \ref{A.A} for this factorization for the multivariate normal probability distribution function case).

\section{Topological regularization with IFN} \label{s.TR}
%As discussed in the introduction, regularization is a methodology used to reduce overfitting by favoring models with a lower number of parameters or with smaller values of them. 
%This is done by penalizing the introduction of extra or large coefficients in a model by modifying the objective function reducing it in proportion to number and/or the weights of the parameters in the model.
%In this paper I use as objective function the log-likelihood of the Student-t distribution and I panelize it with the number of non-zero parameters in the inverse shape matrix $\mathbf{J=  \Omega}^{-1}$. 
%When the penalizing function is linear, this is a case of $L_0$-norm regularization.
The problem is to find the model parameters that maximize likelihood for a given  IFN.
In this paper I address this issue for probabilistic modeling with densities belonging to the elliptical family and I report results for the multivariate normal and Student-t cases. 

For the whole elliptical family the probability density function can be written as:
\begin{equation}\label{e.elliptical}
f(\mathbf X = \mathbf x) = k_p \sqrt{|\mathbf J |} g(\mathbf x, \boldsymbol \mu, \mathbf J),
%\end{equation}
%where, $k_p$ is a constant, $g(\cdot)$ is the so called density generator, a scalar, non-negative Lebesgue measurable function on $[0,\infty)$ such that $\int_0^\infty t^{p/2-1}g(t)dt < \infty$ and  $d_{\mathbf x}^2$ is the following (non-negative) scalar:
%\begin{equation}\label{e.d2}
%d^2 = ({\mathbf x}-{\boldsymbol\mu_{\mathbf X}})^\top\boldsymbol\Omega^{-1}({\mathbf x}-{\boldsymbol\mu_{\mathbf X}}),
\end{equation}
where $\boldsymbol \mu = \mathbb E(\mathbf X)$ are the expected values of $\mathbf X$ and  $\mathbf J$ is a positively defined matrix which coincides with the inverse covariance matrix when it is defined \citep{fang2018symmetric}. %with  $\boldsymbol\Omega$  a positively defined matrix that is proportional to the covariance matrix if the latter exists. 

The matrix  $\mathbf J$ is the quantity I am sparsifying in this proposed  $L_0$-norm topological regularization.
Specifically, only the diagonal $(\mathbf J)_{i,i}$ and elements $(\mathbf J)_{i,j}$ corresponding to edges in the IFN are allowed to be different from zero.  
Therefore, the problem becomes to compute the values of the non-zero elements of $\mathbf J$ that maximize likelihood.
Hereafter, I report the solution for the multivariate normal and Student-t cases.

\subsection{Sparse maximum likelihood solution for the multivariate normal}
The log-likelihood for the multivariate normal distribution is
\begin{definition}[Normal log-likelihood]
Given a set of observations  $\mathbf{\hat x}(s) = (\hat x_1(s),...,\hat x_p(s)) ^\mathsf{T}$ with $s=1...q$,  the log-likelihood of the multivariate normal is
 \begin{equation} 
 \ell(\boldsymbol\mu,\mathbf{J}) =
\frac{q}{2}  \log |\mathbf{ J}  | 
-\frac{1}{2} \sum_{s=1}^q d_{\mathbf{\hat x}(s)}^2
-\frac{qp}{2}\log \pi.
 \label{LLN}\end{equation} 
 \end{definition}
 where
\begin{definition}[Mahalanobis distance]\label{D.Maha}
The term 
\begin{equation}
d_{\hat {\mathbf x}(s)}^2= (\mathbf{\hat x} (s) - \boldsymbol\mu ) ^\mathsf{T} \mathbf{J}   ( \mathbf{\hat x} (s) - \boldsymbol\mu ),
\label{e.Mal}
\end{equation}
is the square of the  Mahalanobis distance \citep{chandra1936generalised}.
\end{definition}
\begin{remark}
The matrix $ \mathbf{J}$ in Eqs.\ref{LLN} and  \ref{e.Mal} must be positively defined and sparse with  non-zero elements only  allowed on the diagonal and or in the off-diagonal positions coinciding with the edges of the associated IFN. 
\end{remark}
%is sparse and it is constructed as.

Now I must find the maximum likelihood solution of Eq.\ref{LLN} under the topological constraint that off-diagonal non-zero elements of $\mathbf{J}$ must coincide with the edges of the given IFN. 
The maximization process is almost identical to the full case but with the topological constraint enforced.
\begin{theorem}[ML solution for $\boldsymbol \mu$ for the sparse multivariate normal problem] 
If $\mathbf J$ is invertible, then the maximum likelihood solution for $\boldsymbol\mu$ %of $\ell(\boldsymbol\mu,\mathbf{J})$ in Eq.\ref{LLN} 
is the sample mean:
 \begin{equation}
 \boldsymbol\mu^* = \frac{1}{q} \sum_{s=1}^q \mathbf{\hat x}(s)
 \end{equation}
\\ 
{\bf Proof}
The proof is identical to the one for the full problem. The maximum of $\ell(\boldsymbol\mu,\mathbf{J})$ in Eq.\ref{LLN}  with respect to $\boldsymbol\mu$ is obtained from the root of
\begin{equation}
\frac{\partial}{\partial \boldsymbol\mu}  \ell(\boldsymbol\mu,\mathbf{J}) =\frac{1}{2} \mathbf J \sum_{s=1}^q \mathbf{\hat x}(s) -\frac{q}{2}  \mathbf J \boldsymbol\mu = 0\;.
\end{equation}
Which is indeed solved by $\boldsymbol\mu^*$ if  $\mathbf J$ is invertible. $\square$
\end{theorem}
The proof that  $\mathbf J$ is invertible is given in lemma \ref{c.J} in  Appendix  \ref{Th.MLNp}.

\begin{theorem}[ML solution for $\mathbf J$ for the sparse multivariate normal problem] \label{Th.MLN}
Given a  IFN structure made of clique and separators, the maximum likelihood solution for the sparse $\mathbf{J}$ is:
\begin{equation}
%{J^*}_{i,j} = \sum_{c\in\mathcal{C}} \left(\hat \mathbf{ \Sigma}_{c}  ^{-1}\right)_{i,j} -  \sum_{s\in\mathcal{S}} \left(\hat \mathbf{  \Sigma}_{s}  ^{-1}\right)_{i,j} .
J^*_{i,j} = \sum_{c\in\mathcal{C}} \left(\hat { \boldsymbol \Sigma}_{c}  ^{-1}\right)_{i,j} -  \sum_{s\in\mathcal{S}} \left(\hat {\boldsymbol  \Sigma}_{s}  ^{-1}\right)_{i,j} ,
\label{J*N}\end{equation}
when $i,j$ belong to a clique of the IFN.
Otherwise ${J^*}_{i,j}=0$ for all other couples of $i,j$ not belonging to cliques.
Where $ \hat { \boldsymbol \Sigma}_{c}$ and $ \hat { \boldsymbol \Sigma}_{s}$ are the Person's sample estimators of the covariances of the variables in the cliques and separators. \\
The proof of this theorem is provided in Appendix \ref{Th.MLNp}.
\end{theorem}
 \begin{remark}
 The sparsification of  $\mathbf{J}$ through Eq.\ref{J*N} provides a way to overcome the curse of dimensionality in the estimation of covariances from observations. Indeed, independently on the overall dimension of the system of variables $\mathbf X$. 
 When the sparse inverse covariance $\mathbf{J}$ is estimated from data, it is then sufficient to have a number of observations, $q$, larger than the size of the largest clique, which is independent from the dimension, $p$, of $\mathbf X$.
 Therefore, through Eq.\ref{J*N} one can obtain well conditioned covariance matrices even when $q \ll p$.
 Equation \ref{J*N} transforms the global problem of estimating the whole matrix inverse into a set of local problems at clique and separator levels. 
 \end{remark}

\subsection{Sparse maximum likelihood solution for the multivariate Student-t} \label{Jlearning}
%The log-likelihood for the multivariate Student-t distribution is
\begin{definition}[Student-t log-likelihood]
Given a set of observations  $\mathbf{\hat x}(s) = (\hat x_1(s),...,\hat x_p(s)) ^\mathsf{T}$ with $s=1...q$,  the log-likelihood of the multivariate Student-t is
 \begin{equation} 
  \ell(\boldsymbol\mu,\mathbf{J} ,\nu) =
q \log \left(\frac{\Gamma\left(\frac{\nu+p}{2}\right)}{(\nu-2)^{p /2}\pi^{p /2}\Gamma(\frac\nu2)} \right)
+ \frac{q}{2}  \log|\mathbf{ J}  | 
-\frac{\nu+p }{2} \sum_{s=1}^q \log \left( 1+ \frac{1}{\nu-2}  d^2_{\hat {\mathbf x}(s)}  \right).
 \label{LLS}\end{equation} 
 \end{definition}
 Where $d^2_{\hat {\mathbf x}(s)}$ is the square Mahalanobis distance as defined in definition \ref{D.Maha}; $\mathbf J$ is the inverse covariance and $\nu$ is the degrees of freedom that here we assume being always larger than 2. 
 Indeed, for $\nu \le 2$ the covariance is not defined. 

In the non-sparse (full) case it is known that the likelihood of multivariate Student-t models can be maximized by means of a procedure known as expectation-maximization (EM) introduced by \cite{dempster1977maximum} (see also \cite{bishop2006pattern} Chap.9).

I shall show hereafter that such a procedure can be applied  also to the maximization the likelihood of the sparse Student-t model for any given chordal IFN structure. 

\begin{theorem}[ML solution for $\boldsymbol \mu$ for the sparse multivariate Student-t problem] \label{Th.Smu}
If $\mathbf J$ is invertible, then the maximum likelihood solution for $\boldsymbol\mu$ is a weighted mean:
 \begin{equation}\label{e.StEMmu}
 \boldsymbol\mu^* = \frac{1}{ \sum_{s=1}^q w_s^*} \sum_{s=1}^q w_s^* \mathbf{\hat x}(s) ,
 \end{equation}
with weights
\begin{equation}\label{e.StEMweights}
w_s^* = \frac{\nu+p}{{\nu+\frac{\nu}{\nu-2}d^{*2}_{\hat {\mathbf x}(s)}}}.
\end{equation}
Proof is provided in Appendix \ref{s.ST}.
\end{theorem}
The $d^2_{\hat {\mathbf x}(s)}$ is the Mahalanobis distance computed using the ML solution $\mathbf J^*$ (see Theorem \ref{Th.JMLSt}).
The $w_s^*$ are the asymptotic solutions for $t\to \infty$ of the recursive EM process.

The ML solution for the sparse $\mathbf J$ is also obtained with the EM approach.
\begin{theorem}[ML solution for $\mathbf J$ for the sparse multivariate Student-t problem] \label{Th.JMLSt}
The maximum likelihood solution for $\mathbf{J}$  is:
\begin{equation}
%{J^*}_{i,j} = \sum_{c\in\mathcal{C}} \left(\hat \mathbf{ \Sigma}_{c}  ^{-1}\right)_{i,j} -  \sum_{s\in\mathcal{S}} \left(\hat \mathbf{  \Sigma}_{s}  ^{-1}\right)_{i,j} .
J^*_{i,j} = \sum_{c\in\mathcal{C}} \left( { \boldsymbol \Sigma}_{c}^{*-1}\right)_{i,j} -  \sum_{s\in\mathcal{S}} \left( {\boldsymbol  \Sigma}_{s} ^{*-1}\right)_{i,j} ,
\label{J*S}
\end{equation}
when $i,j$ are an edge of a clique.
Otherwise ${J^*}_{i,j}=0$ for all other couples of $i,j$ not belonging to cliques.
Where $ { \boldsymbol \Sigma}^*_{c}$ and $  { \boldsymbol \Sigma}^*_{s}$ are the EM estimators of the covariances of the variables in the cliques and separators, which are given by the weighted sample averages:
\begin{equation}
 { \boldsymbol \Sigma}^*_{i,j}
=  \!\frac1q {\sum_{s=1}^q w_s^* (\hat { x_i}(s) - \mu_i^*)^\top(\hat { x_j}(s) - \mu_j^*)}.
\label{S*}
\end{equation}
\\
The proof of this theorem is provided in Appendix \ref{s.ST}.
\end{theorem}

The sparsity of $\mathbf J$ is not affecting the form of the EM solutions which have the same form also in the full case. However, in the sparse case only the elements belonging to cliques must be computed which reduces computational complexity from  $\mathcal O(p^2)$ to $\mathcal O(p)$.

The parameter $\nu$ can also be computed through the EM procedure. However, I prefer to estimate it  independently by estimating via a power law fit  of the left and right tails of the probability distribution of all the univariate marginals of $\mathbf X$. 
Indeed, all marginal Student-t distributions of $\mathbf X$ must behave as a power law on both left and right tails  with tail-exponent $\nu$.

\section{Experiments}\label{s.Ex}
In order to test the novel topological regularization methodology introduced with this paper I computed and compared the likelihood  of several models using three kind of data.
\subsection{Data}
I collected daily prices from $623$ stocks continuously traded on the US equity market between 01/02/1999 and 20/03/2020 for a total of 5515 trading days.
For each stock  `$i$' ($=1,...,p$) I computed the log-returns, $\hat x_i(s) = \log Price_i(s) - \log Price_i(s-1)$, ($s=1,...,q)$.
Results are computed over 100  random re-sampled datasets generated by randomly picking with repetitions $p=100$ different return series among the 623 stocks. 
For each random choice of the $p=100$ return series I randomly sampled $2\times q$ returns without repetition using $q$ observations for the training set and  $q$ observations for the test set. % and $q$ observations for the test set. 
I performed two sets of experiments with $q=150$ and $q=600$ respectively.   
I also tested the procedure on synthetic datasets  artificially generated from multivariate normal distributions and multivariate Student-t distributions. In these cases I used the empirical covariance and means from the real data as parameters to generate artificial datasets with properties consistent with the real data. The Student-t was generated with $\nu=2.2$ degrees of freedom. 
Analogously with the real data I generated 100  random datasets of $p=100$ multivariate variables. 
I used $q=600$ observations for the training set and also $q=600$ observations for the test set. %and $q$ observations for the test set.

\subsection{Model construction and parameter estimation}
For each training dataset I generated MFCF networks with maximum clique sizes in the range from 2 to 100. 
For each maximum clique size I generate two different networks by using the Pearson correlation estimate and the Kendall correlation estimate. 
As MFCF gain function I chose the sum of the squares correlations, which is one of the simplest choices that produces cliques all of sizes equal to the maximum clique size. 
The MFCF networks I generate have separators that are used only once (multiplicity one).  
As degrees of freedom I empirically investigated the tails of the marginal distributions across the whole dataset retrieving a tail exponent $\nu = 2.2$ as a good average estimator for the degrees of freedom. 
I verified that relative results are little sensitive to this parameter although the values the likelihood can change sensibly with $\nu$.
The  covariances are retrieved by multiplying by the standard deviations the elements of the  correlations matrices. 
Using these MFCF networks I then compute the maximum likelihood inverse sparse covariance estimates for multivariate normal modeling, as described in Eq.\ref{J*N}, and for the multivariate Student-t modeling, as described in Eq.\ref{J*S}. 

\subsection{Comparison with GLasso}
In order to compare the results with a meaningful state-of-the-art sparse modeling approach, I computed $L_1$-norm regularized sparse inverse covariance estimators by using a Quadratic Approximation for Sparse Inverse Covariance Estimation (QUIC) by \cite{hsieh2014quic}.\footnote{Matlab implementation available at: {http://www.cs.utexas.edu/~sustik/QUIC/.}}
Different levels of sparsity were achieved by varying the regularization penalty, $\lambda$, with values between  $10^{-6}$ and $10^{-3}$.
%For this purpose I used the QUICK package in matlab \cite{•}{QUICK} 
%In analogy with the experiments with MFCF, I resamples 
%first computed the Kendall's correlation and from it I generated a set of IFN graphs using the MFCF algorithm with maximum clique sizes in the range from 2 to 100. 
%As MFCF gain function I chose the sum of the squares correlations, which is one of the simplest choices that produces cliques all of sizes equal to the maximum clique size. 
%%I then computed the sparse inverse shape matrix on the training set using Eq.\ref{J} from the Kendall correlation coefficients multiplied by the standard deviations and by the factor $(\nu-2)/\nu$.
%The covariances are retrieved by multiplying by the standard deviations the elements of the Kendall correlation. 
%As degrees of freedom I empirically investigated the tails of the marginal distributions across the whole dataset retrieving a tail exponent $\nu = 2.2$ as a good average estimator for the degrees of freedom. I verified that results are qualitatively little sensitive to this parameter although quantitatively the likelihood changes sensibly with $\nu$.
%I optimized the parameters to maximize likelihood on the train set by using the EM procedure described in section \ref{Jlearning}.

\subsection{Results}
I computed the mean log-likelihood  $\ell$  for the range of MFCFs with different clique sizes and for both multivariate normal and multivariate Student-t models computed by using either the Pearson's and the Kendall's covariance estimators and the Expectation Maximization procedure  (see \ref{Jlearning}).
The largest mean log-likelihoods across the MFCF clique-sizes' range and the value of the corresponding clique size are reported in Table \ref{t.Validation} for all the models.
The parameters are estimated on the training set and the results are instead reported for the test set. 
One can observe that for real data the sparse Student-t model constructed by using Kendall's covariance and Expectation Maximization procedure gives the best results for both $q=150$ and $600$ with smaller clique size selected for the shorter dataset. The combination Student-t model, Kendall's covariance and Expectation Maximization procedure is also best for the multivariate Student-t synthetic datasets. 
Conversely, for the multivariate Normal synthetic datasets the best results are achieved by the sparse Normal model construct using Peterson's covariance.
\begin{table}
\begin{tabular}{ c } 
Max average log-likelihood per observations $\ell/q$; and max clique size \\
\begin{tabular}{ | r |  lr  | lr | lr | lr |} 
%\begin{tabular}{l} Real data\\ $q=150$ \end{tabular} & \begin{tabular}{c}  \\ $600$ \end{tabular}&  
%\begin{tabular}{l}Normal data\\ $q=150$\end{tabular}& \begin{tabular}{c}  \\ $600$\end{tabular}& 
%\begin{tabular}{l}Student-t data \\ $q=150$\end{tabular}& \begin{tabular}{c}  \\ $600$\end{tabular}
\hline
Estimator & Real q=150 && Real q=600 && St-t. q=600 && Nor. q=600 &\\
\hline 
%Normal Sparse Model from Pearson's Covariance                   	
Nor.; Per. & 350.9; & 5    		&362.6; &10  			& 344.0; & 5   	&  {\bf 371.3;} &30  \\ 
Nor.; Ken. & 360.4; & 20    	&363.8; &100  			& 360.6; & 100  &   369.4; &100  \\ 
St-t.; Per. & 376.7; & 6    		&383.3; &11 			& 447.0; & 7   	&   363.3; & 30  \\ 
St-t.;  Per. EM &  383.9;&  8   	&389.6; &20  			& 459.6; & 50   	&   366.8; & 30  \\ 
St-t.;  Ken. & 381.0; & 15    	&385.6; & 50  			& 454.7; & 100   &   364.8; & 100  \\ 
St-t.;  Ken. EM & {\bf 384.8;} & 8    &{\bf 389.8}; & 15  	& {\bf 460.0;} & 30  &366.8; & 30  \\ 
\hline
\end{tabular}
\end{tabular} 
\caption{
Summary of results for the maximum values in the test set for the mean likelihood $\ell/q$ per observation and the corresponding  max-clique-size in the MFCF network.
Several modes are investigated: Normal likelihood  (Nor., see Eq.\ref{LLN}), Student-t likelihood  (Nor., see Eq.\ref{LLS}), Pearson covariance estimator (Per.), Kendall covariance estimator (Ken.), expectation maximization parameters estimation (EM, see Eqs.\ref{e.StEMmu}, \ref{J*S}).
%Summary of results for the mean likelihood $\ell/q$ (Eq.\ref{LLS}), mean penalized likelihood $\mathcal L/q = \ell/q - \left\lVert \mathbf{J}  \right\rVert_0$ and IFN-MFCF max-clique-size computed on the test set for the MFCF, TMFG and FULL models.
%The MFCF are sparse models with IFN network learned on the train set, parameters  optimized on the train set and max clique size selected on the validation set.   
%The TMFG is also learned and optimized on the train set.
%The FULL is a non-sparse model optimized on the train set.
%Results are means computed on the test set over 100 different re-samplings for the real data and new realizations for the artificial data.
%Columns respectively refer to: real log-return data (Real data) , multivariate normal synthetic data (Normal data), multivariate Student-t synthetic data (Student-t data); all for both  $q=150$ and $600$.
\label{t.Validation}
}
\end{table}

\begin{figure}
\begin{center}
\includegraphics[width=0.7\textwidth]{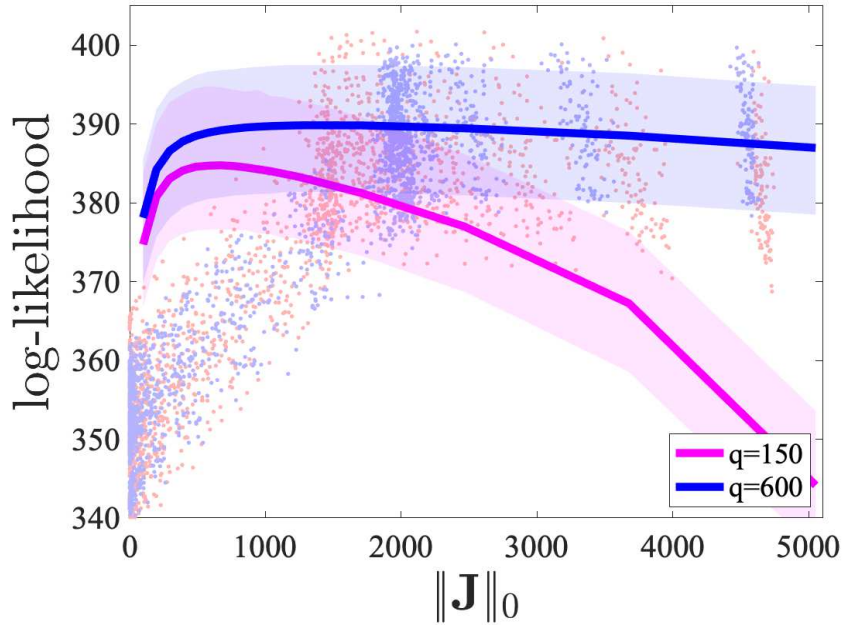}
\caption{
Log-likelihood $\ell(\boldsymbol\mu,\mathbf{J} ,\nu)/q$ (Eq.\ref{LLS})  for Student-t models with sparse inverse covariance matrix $\mathbf{J}$ constructed by using the MFCF approach with IFN graphs with different levels of sparsity obtained by changing the maximum clique size from 2 to 100.
The IFN have been constructed using the Kendall estimate of the correlation matrix.
The x-axis reports $\left\lVert \mathbf{J}  \right\rVert_0$ which is the number of edges in the IFN graph.
Models have been optimized to maximize Student-t likelihood by using the EM procedure described in section \ref{Jlearning}. 
Parameters are estimated on two train sets of `real' data (see text) with different lengths: $q=150$ and $q=600$ respectively (blue and magenta).
Reported results are the log-likelihoods computed on the test set.
The lines are the means and the bands around the lines are the 10\% and 90\% quantiles over 100 random re-sampling. 
The points are instead from Glasso models computed using  QUIC Quadratic Approximation for Sparse Inverse Covariance Estimation implemented by  \cite{hsieh2014quic} using a range of regularization penalty between  $10^{-6}$ to $10^{-3}$.
%http://www.cs.utexas.edu/~sustik/QUIC/.
\label{f.TrainValidationResults}}
\end{center}
\end{figure}

Fig. \ref{f.TrainValidationResults} reports results for the Student-t log-likelihood (Eq.\ref{LLS}) estimated using Kendall covariance and expectation maximization. 
The parameters are estimated on the training set and the results are instead reported for the test set.  The log-likelihood is computed for a range of sparsity values obtained by varying the m,aximum clique size from 2 to 100 (complete graph network). 
The x-axis reports the number of edges in MFCF  (i.e. the number of non-zero elements in the sparse inverse covariance $\left\lVert \mathbf{J}  \right\rVert_0$).
The tick lines are averages over 100 re-samplings and the bands are the $10\%$ and $90\%$ quantiles. Note that, the re-sampling picks randomly both the time series and the returns. Therefore the observed consistency and the relatively narrow quantile band are strong indications of statistical robustness of the results.
Also note that the last points on the right of the two plots are the full models (max clique = 100) with the complete (non-sparse) inverse covariance matrix.
As one can see, for small observation sets ($q=150$) the sparse models largely over-perform  the complete models. 
Whereas, for larger observation sets ($q=600$) the difference is smaller.

In the figure, I also report for comparison results obtained by estimating sparse inverse covariance via $L_1$-norm regularization using the QUIC package by \cite{hsieh2014quic}.
One can see that for sparse modeling, up to $\left\lVert \mathbf{J}  \right\rVert_0 \sim 10 \times p = 1,000$ the MFCF approach is largely over-performing the QUIC results. 
For denser networks (i.e. $\left\lVert \mathbf{J}  \right\rVert_0 \sim 2,000$) the MFCF and  QUIC approach deliver similar results. For instance, for $q=600$, the QUIC approach with $\lambda =  2\,10^{-5}$ retrieves 1,720 average number of edges and average $\ell/q =  253.9$ with $[248.7,258.6]$ the 10\% and 90\% quantiles.
By comparison, the MFCF for max clique equal to 20 has 1,710 edges and average likelihood $\ell/q =  254.0$ with quantiles $[248.9,258.7]$.
Similar results are retrieved for other levels of sparsity with the QUIC results slightly improving performances over the MFCF when the model becomes denser.

Other results with different combination of models and with artificial data are reported in Appendix \ref{A.Comparison}.
Specifically, I test normal modeling on the real datasets (Fig.\ref{f.RealNormal}) and I test both normal and Student-t models on synthetic datasets produced with normal and Student-t  distributions (Figs.\ref{f.SynteticNormal} and \ref{f.SynteticStudentTl}).
Overall, I observe a very consistent picture across al experiments and the various combinations of model construction and data. 
It results that Student-t modeling is more appropriate for the real financial data resulting in larger likelihoods.
Not surprisingly, it results that normal models works better on normal data and instead Student-t models have higher likelihoods on Student-t data.
The construction with Kendall's estimate of the covariance is producing better results for the real data and the Student-t synthetic data but not for the multivariate normal synthetic data where the Pearson's estimate is better. 
The expectation-maximization optimization procedure used for the Student-t models makes the difference between Kendall's  and Pearson's estimates small, but still quantifiable. 
Indeed, EM can `cure' the parameters estimate and therefore it is little sensitive to the starting matrix, however the IFM networks from Kendall's or Pearson's estimates are not identical and this produces the difference.  
Consistently with what I reported for the real data with Student-t modeling (Fig.\ref{f.TrainValidationResults}),  Glasso models underperform for all sparse networks and then achieve comparable performances to the IFM-LoGo models at higher levels of network density (above $\left\lVert \mathbf{J}  \right\rVert_0 \sim 2,000$) which correspond to rather dense networks with about 40\% of edges present.

\section{Conclusions and perspectives}\label{s.C}
In this paper I have introduced a methodology for topological $L_0$-norm regularization with information filtering networks and I have applied it in detail for a case of penalized likelihood in Student-t sparse modeling.
The regularization methodology consists in  keeping different from zero only the parameters of the multivariate distribution that correspond to  edges in the IFN.
By using clique forests IFNs, one guarantees positive definiteness and decomposition into local parts of the inverse covariance matrix (Eqs.\ref{J*N} and \ref{J*S}).
This is an important property associated with this kind of IFN and it applies to the vast class of models belonging to the elliptical family \citep{fang2018symmetric} which includes the Student-t but also the normal, the Laplace and the multivariate stable distributions. 
It is actually more general than this, applying also to non symmetric multivariate distributions such as the generalized hyperbolic family.  
This  $L_0$-norm topological regularization methodology strongly improves model interpretability because IFN structures are known to meaningfully represent relevant interrelations in complex data structures with a vast literature reporting their successful applications to various domains from finance to biology.

I have shown that the expectation-maximization methodology commonly used to estimate the maximum likelihood coefficients in  Student-t models can be used also for this $L_0$-norm regularized sparse models with the advantage that in this case computation must be done only for the coefficients corresponding to IFN edges reducing computation complexity from $\mathcal O(p^2)$ to $\mathcal O(p)$.

Experiments on real datasets from  equity prices and multivariate synthetic datasets demonstrate that the proposed methodology is directly applicable to a range of practically relevant problems. Results demonstrate that  $L_0$-norm topologically  regularized models outperform the full models and reveal that smaller observation sets selects sparser IFN models. 
A comparison with $L_1$-norm regularization by Glasso approach, shows that the proposed methodology largely outperforms Glasso for sparse models and tend to perform similarly for denser models.  Furthermore, if must be noticed that the proposed IFN-LoGo approach is computationally more efficient and the sparse network has better interpretability.  

The present paper reports exclusively on $L_0$-norm regularization via IFN priors, however, the nature of this sparsification allows to combine straightforwardly $L_1$ and $L_2$ regularizations as well within this methodology. 
Indeed, Theorems \ref{Th.MLN} and \ref{Th.JMLSt} provide a formula for the maximum likelihood solution of the sparse inverse covariance matrix as sum of local inverse matrices associated with the clique and separator sets. 
On such local inversions, shrinkage and lasso regularization can be applied directly. 
This has the further advantage that both the inversions and the regularizations are on local-small dimensional matrices making the procedure computationally efficient and fully parallelizable.

%I think it is important to stress that, although I have focused  the illustration of this topological regularization methodology to the very specific case of likelihood maximization for a multivariate Student-t model, the method itself is  general.
%For instance, Eq.\ref{MY|X} provides the expression for the $L_0$-norm IFN penalized multilinear regression for the elliptical family. 
%However, this will be the subject of future works.

\section*{Acknowledgmets}
The author acknowledges discussions with several members of the Financial Computing and Analytics group at UCL. A special thank to Dr. Guido Massara for many critical inputs and to Killian Guillaume Paul Martin-Horgassan for discussions. 
Also, thanks for support from ESRC (ES/K002309/1),  EPSRC (EP/P031730/1) and EC (H2020-ICT-2018-2 825215).

\appendix
\vskip2cm
{\Large\bf\raggedright Appendix}

\section{Decompositions for the multivariate normal case} \label{A.A}
%This is a direct consequence of Eq.\ref{MultiVarNormalDecomp} and the proof is provided in  \cite{lauritzen1996}.
\begin{theorem}[Decomposition of the sparse multivariate normal distribution]\label{Th.DecJstruct}
Given a sparse $\mathbf J$ inverse covariance with a chordal IFN structure where the non-zero entries corresponds to a set of cliques $\mathcal C$ and separators $\mathcal S$ in a clique-forest, the sparse multivariate normal probability density function, $\varphi(\mathbf{X=x} | \boldsymbol\mu,\mathbf{ \Sigma})$, can be decomposed in terms of cliques and separators as follows: 
\begin{equation}
\varphi(\mathbf{X=x} | \boldsymbol\mu,\mathbf{ \Sigma})
= 
\frac{\prod_{c\in \mathcal{C}} \varphi(\mathbf{X_c=x_c} | \boldsymbol\mu_c,\mathbf{ \Sigma_c}) }
{\prod_{s\in \mathcal{S}}  \varphi(\mathbf{X_s=x_s} | \boldsymbol\mu_s,\mathbf{ \Sigma_s})} \;.
\label{MultiVarNormalDecomp}
\end{equation}
\\ \noindent
{\bf Proof}
The proof is a straightforward  consequence of the exponential form of the normal distribution and it is for instance provided in  \cite{lauritzen1996}.
$\square$
\end{theorem}

\begin{theorem}[Decomposition of conditionally independent multivariate normal variables] \label{Th.DecCondProb}
Given a set of multivariate normal variables corresponding to a set of cliques $\mathcal C$ which are conditionally independent from each other when conditioned to their separators $\mathcal S$ in a clique-forest structure, then the multivariate normal probability density function can be decomposed in terms of cliques and separators as follows: 
\begin{equation}
\varphi(\mathbf{X=x} | \boldsymbol\mu,\mathbf{ \Sigma})
= 
\frac{\prod_{c\in \mathcal{C}} \varphi(\mathbf{X_c=x_c} | \boldsymbol\mu_c,\mathbf{ \Sigma_c}) }
{\prod_{s\in \mathcal{S}}  \varphi(\mathbf{X_s=x_s} | \boldsymbol\mu_s,\mathbf{ \Sigma_s})} \;.
\label{MultiVarNormalDecomp2}
\nonumber \tag{\ref{MultiVarNormalDecomp}}
\end{equation}
\\ \noindent
{\bf Proof}
The proof is a direct consequence of the Bayes formula and the proof is provided in  \cite{lauritzen1996}.
$\square$
\end{theorem}

It is clear that the two formulas in theorems \ref{Th.DecJstruct} and \ref{Th.DecCondProb} are the same (indeed they have the same number \ref{MultiVarNormalDecomp}), however they are consequences of two different facts that happen to coincide for the multivariate normal probability density function. 

\begin{remark}
%It must be noted that this decomposition implies and is a consequence of the independence between variables associated with the cliques in the IFN when conditioned to the separators. 
The conditional independence is an exclusive property of the sparse multivariate normal and it is not applicable for the Student-t case.
\end{remark}

As a consequence of Eq.\ref{MultiVarNormalDecomp} one has that the non-zero elements of the sparse covariance matrix $\mathbf J$ can be expressed as a simple sum of local inverse covariances.

\begin{corollary}[Decomposition of the inverse covariance matrix]\label{c.Jxx}
The elemets of the inverse covariance are given by:
\begin{equation}
J_{i,j} = \sum_{c\in\mathcal{C}} \left( { \boldsymbol \Sigma}_{c}  ^{-1}\right)_{i,j} -  \sum_{s\in\mathcal{S}} \left( {\boldsymbol  \Sigma}_{s}  ^{-1}\right)_{i,j},
\label{J*Npxx}\end{equation}
\\ \noindent
{\bf Proof}
This is a direct consequence of Eq.\ref{MultiVarNormalDecomp} and the proof is provided in  \cite{lauritzen1996}. $\square$
\end{corollary}

There are other two useful consequences of the decomposition in Eq.\ref{MultiVarNormalDecomp}.

\begin{corollary}[Decomposition of the determinant]\label{c.J1}
\begin{equation}
|\mathbf J| =  \frac{\prod_{c\in \mathcal{C}}  |\mathbf J_c|}{\prod_{s\in \mathcal{S}}  |\mathbf J_s|} .
\label{LoGoDetreminants}\end{equation}
\\ \noindent   
{\bf Proof}
This is a direct consequence of Eq.\ref{MultiVarNormalDecomp} and the proof is provided in  \cite{lauritzen1996}.
$\square$
\end{corollary}

\begin{corollary}[Decomposition of the Mahalanobis distance]\label{c.J2}
\begin{equation}
d^2 =  \sum_{c\in \mathcal{C}}  d^2_c - \sum_{s \in \mathcal{S}}  d^2_s
\label{LoGoMahanobis}\end{equation}
with $d^2_c=(\mathbf{x} - \boldsymbol\mu_c) ^\mathsf{T}\mathbf{J}_c(\mathbf{x} - \boldsymbol\mu_c)$ and 
$d^2_s=(\mathbf{x} - \boldsymbol\mu_s) ^\mathsf{T}\mathbf{J}_s(\mathbf{x} - \boldsymbol\mu_s)$.
\\ \noindent   
{\bf Proof}
This is a direct consequence of Eq.\ref{MultiVarNormalDecomp} and the proof is provided in  \cite{lauritzen1996}.
$\square$
\end{corollary}

\section{Theorems and proofs for normal ML}
\begin{lemma}[Positive definitness]\label{c.J}
The sparse inverse covariance $\mathbf{J}$ constructed from Eq.\ref{J*N} is positively defined if $\mathbf{  \Sigma}_{c}$ and $\mathbf{   \Sigma}_{s}$ are positively defined. 
\\ 
{\bf Proof}
A sum of positively defined matrices is positively defined. $\square$
\end{lemma}

\noindent
{\bf Proof} ({\bf Proof of Theorem \ref{Th.MLN}}) \label{Th.MLNp}
\\
I have to prove that the sparse inverse covariance matrix constructed using Eq.\ref{J*N} is the maximum likelihood solution for the sparse multivariate normal case for a give IFN sparsity structure.\\
{\it I develop this proof into two steps.
\begin{itemize} 
\item[1.] {\bf First}, I show that when $i,j$ is an edge of a clique of the IFN structure then the solution for the covariance coefficient must be the Person's sample covariance estimator between variable $i$ and variable $j$.
This part proceed in the same way as for the full problem. %but sparsity constraint must be accounted for. 
In particular, the maximum of $\ell(\boldsymbol\mu,\mathbf{J})$  with respect to $\mathbf J$ is obtained from the root of
\begin{equation}
\frac{\partial}{\partial J_{i,j}}  \ell(\boldsymbol\mu,\mathbf{J}) =
\left. \frac{q}{2} (\mathbf J^{-1})_{i,j} -\frac{1}{2} \sum_{s=1}^q  ({\hat x}_{i}(s) - \mu_{i})({\hat x}_{j}(s) - \mu_{j})
\right|_{\mathbf J = \mathbf J^*} = 0\;,
\label{e.LLNmax}
\end{equation}
for $(i,j) \in c$. 
This therefore implies that the elements $i,j$ in the maximum likelihood covariance must coincide with the Person's sample covariance estimator, $(\hat { \boldsymbol \Sigma})_{i,j}$, when the couple $i,j$ is an edge of a clique. 
%Conversely, one must keep $J_{i,j}=0$ when $i,j$ is not an edge of a clique accordingly with the topological constraint.
\item[2.] {\bf Second},I demonstrate that the sparsity structure of $\mathbf J$ over a chordal graph imposes that the inverse covariance must be in the form
\begin{equation}
%{J^*}_{i,j} = \sum_{c\in\mathcal{C}} \left(\hat \mathbf{ \Sigma}_{c}  ^{-1}\right)_{i,j} -  \sum_{s\in\mathcal{S}} \left(\hat \mathbf{  \Sigma}_{s}  ^{-1}\right)_{i,j} .
J_{i,j} = \sum_{c\in\mathcal{C}} \left( { \boldsymbol \Sigma}_{c}  ^{-1}\right)_{i,j} -  \sum_{s\in\mathcal{S}} \left( {\boldsymbol  \Sigma}_{s}^{-1}\right)_{i,j},
\label{J*Np1}\end{equation}
where ${ \boldsymbol \Sigma}_{s} $ and ${ \boldsymbol \Sigma}_{s} $ are respectively the covariances  of the  distributions of the subsets of variables in the cliques and separators. 
This is a direct consequence of the decomposition property for the multivariate normal distribution (see Eq.\ref{MultiVarNormalDecomp}).
%\item[3.] Finally, I must demonstrate that when in Eq.\ref{J*Np1}, ${ \boldsymbol \Sigma}_{c}=\hat { \boldsymbol \Sigma}_{c}$ and ${ \boldsymbol \Sigma}_{s}=\hat { \boldsymbol \Sigma}_{s}$ then $(\mathbf J^{-1})_{i,j} =(\hat { \boldsymbol \Sigma})_{i,j}$.
%This is direct consequence of the fact that the distributions of the subsets of variables in the cliques and separators are multivariate normals with ${ \boldsymbol \Sigma}_{c}=\hat { \boldsymbol \Sigma}_{c}$ and ${ \boldsymbol \Sigma}_{s}=\hat { \boldsymbol \Sigma}_{s}$ as covariances.
\end{itemize}
As a consequence, the ML sparse inverse covariance solution must have the form of Eq.\ref{J*Np1} with elements  given by the sample covariances, and this is indeed Eq.\ref{J*N}.
$\square$
}

\section{ML solution for the Student-t distribution } \label{s.ST}
Let me start from the definition of the multivariate Student-t probability density function.

\begin{definition}[Multivariate Student-t distribution]
Given of a set of random variables $\mathbf{X}\in \mathbb{R}^{p \times 1}$ the multivariate Student-t probability density function has the following canonical general expression \citep{kotz2004multivariate}: 
\begin{equation}
t(\mathbf{X=x})= 
\sqrt{\frac{1}{  |{\mathbf{  \Omega}}| (\nu\pi)^{p}}}
\frac{\Gamma\left(\frac{\nu+p}{2}\right)}{\Gamma(\frac\nu2)} 
\left(
1+\frac{ ({\mathbf x}-{\boldsymbol\mu}) ^\mathsf{T}  \mathbf{  \Omega}^{-1} ({\mathbf x-{\boldsymbol\mu})}}{\nu} 
\right)^{\!\!\!-\frac{\nu+p}{2}}
\label{p.Z}\end{equation}
where  $\boldsymbol\mu \in \mathbb{R}^{p\times 1}$ is the vector of location parameters;  $\mathbf{ \Omega} \in \mathbb{R}^{p \times p}$ is a positively defined matrix called  shape matrix; and $\nu > 0$ is a scalar called degrees of freedom.
\end{definition}
The covariance matrix is defined when $\nu>2$ and it is given by
$$
\boldsymbol \Sigma = \frac{\nu}{\nu-2} \boldsymbol \Omega.
$$
Assuming, $\nu>2$, consistently with the previous notation for the normal case I re-write the expression for the Student-t distribution in terms of the inverse covariance matrix $\mathbf J$.
\begin{equation}\label{e.mySTdef}
t(\mathbf{X=x})= 
\sqrt{\frac{|\mathbf J|}{ ((\nu-2)\pi)^{p}}}
\frac{\Gamma\left(\frac{\nu+p}{2}\right)}{\Gamma(\frac\nu2)} 
\left(
1+\frac{ ({\mathbf x}-{\boldsymbol\mu}) ^\mathsf{T}  \mathbf J ({\mathbf x-{\boldsymbol\mu})}}{\nu-2} 
\right)^{\!\!\!-\frac{\nu+p}{2}}
\end{equation}

The EM construction makes use of the fact that the multivariate Student-t can be written as a normal mixture representation: 
 \begin{equation}
% t(\mathbf{X=x} | \boldsymbol\mu,\mathbf{ \Sigma},\nu ) = \int  \varphi\left(\mathbf{x} | \boldsymbol\mu,\mathbf{ \Sigma}/z \right) \pi( z | \nu ) dz\;.
t(\mathbf X= \mathbf x)  = \int_{0}^{+\infty}  h(z | \frac{\nu}{2},\frac{\nu}{2})  \varphi(\mathbf x | \boldsymbol \mu,\frac{\nu}{\nu-2}\mathbf J z) d z
\label{NormalMixture}\end{equation}
Where  
\begin{equation} \label{e.N}
\varphi\left(\mathbf{x} | \boldsymbol\mu,\frac{\nu}{\nu-2} \mathbf J z \right) = 
 \sqrt{\frac{{z^p \nu^p |\mathbf J|}}{{(2\pi(\nu-2))^{p}}}}
\exp{\left[ -\frac{z}{2}\frac{\nu}{\nu-2}
(\mathbf{x} - \boldsymbol\mu)^\top \mathbf J (\mathbf{x} - \boldsymbol\mu) \right]},
\end{equation}
is the multivariate normal density function with $\boldsymbol\mu\in \mathbb{R}^{p \times 1}$  the location parameters and $\mathbf J z \in \mathbb{R}^{p \times p}$ is a rescaled covariance matrix. 
Instead $h(z | \frac{\nu}{2},\frac{\nu}{2})$ is  the probability density function of a gamma distribution 
\begin{equation} \label{e.G}
h(z |\alpha,\beta) = \frac{\beta^\alpha}{\Gamma(\alpha)} z^{\alpha - 1} e^{-\beta z } \;\;.
\end{equation}
with both scale $\alpha$ and rate $\beta$ parameters equal to $\frac{\nu}{2}$.   

Let me then recap the expectation maximization (EM) approach step by step, explicitly taking into account the sparsity of $\mathbf J$ in our case.
The EM approach proceeds into two main steps. 
The E-step, where and expectation function is defined; then an M-step, where it is maximized recursively.

%%%%%%%%%%%%<<<<<
Let me call
\begin{equation}\label{e.f}
f(\mathbf x , z | \boldsymbol \mu, {\mathbf J}, \nu ) = h(z | \frac{\nu}{2},\frac{\nu}{2})  \varphi(\mathbf x  | \boldsymbol \mu, \frac{\nu}{\nu-2}{\mathbf J} z).
\end{equation}

\begin{itemize}
\item  {\bf E step.}\\
I define the following expectation:
\begin{equation}%\label{e.EstepE1}
Q(\boldsymbol \mu, {\mathbf J} | \boldsymbol \mu^t, {\mathbf J}^t) 
= \sum_{s=1}^q \int_0^\infty 
 f( z | \hat {\mathbf x}(s), \boldsymbol \mu^t, {\mathbf J}^t, \nu  )  \log  f({\hat {\mathbf x}(s) , z } |\boldsymbol \mu, {\mathbf J}, \nu)  d z .
\end{equation}

\item {\bf M step.} 
I now search for the maxima of the expectation by differentiating with respect the parameters and equalling to zero.
\begin{equation}%\label{e.MstepE1}
\nonumber
\frac{\partial}{ \partial \boldsymbol \mu } 
Q(\boldsymbol \mu, {\mathbf J} | \boldsymbol \mu^t, {\mathbf J}^t)  
 = \left.  \frac{\nu}{\nu-2}\sum_{s=1}^q \!\! 
\int_0^\infty \!\! z f( z | \hat {\mathbf x}(s), \boldsymbol \mu^t, {\mathbf J}^t, \nu  )  
(\hat {\mathbf{x}}_s - \boldsymbol\mu)^\top \mathbf J dz \right|_{\boldsymbol \mu =\boldsymbol \mu^{t+1}} \!\! \!\!\!\! \!= \! 0 
\end{equation}
which, if $\mathbf J$ is positively defined, results in the solution
\begin{equation}\label{e.Mut1}
\boldsymbol \mu^{t+1} =  \frac{\sum_{s=1}^q w_s^t  \hat{\mathbf x}(s) }{\sum_{s=1}^q w_s^t},
\end{equation}
with
\begin{equation}\label{e.wkt}
w_s^t =  \int_0^\infty z  f( z | \hat {\mathbf x}(s), \boldsymbol \mu^t, {\mathbf J}^t, \nu  ) dz
\end{equation}
which can be computed explicitly. 
Indeed, substituting Eq.\ref{e.N} and \ref{e.G} one has 
\begin{align} \nonumber
w_s^t \propto \int_0^\infty z g(z | \frac{\nu+p}{2},\frac{\nu+\frac{\nu}{\nu-2} d^2_{\hat {\mathbf x}(s)}}{2} dz.
\end{align}
that is the expected value for a gamma distribution with $\alpha = \frac{\nu+p}{2}$ and 
$\beta = \frac{\nu+\frac{\nu}{\nu-2}d^2_{\hat {\mathbf x}(s)}}{2}$ which is
\begin{equation}\label{e.StEMweights1}
w_s^t = \frac{\nu+p}{{\nu+\frac{\nu}{\nu-2}d^2_{\hat {\mathbf x}(s)}}}.
\end{equation}
where the quantity
\begin{equation}
d^2_{\hat {\mathbf x}(s)} = (\mathbf{\hat x} (s) - \boldsymbol\mu^t ) ^\mathsf{T} \mathbf{J}^t   ( \mathbf{\hat x} (s) - \boldsymbol\mu^t ),
\end{equation}
depends on the stage $t$ of the EM process and therefore  $w_s^t$ must be computed recursively.
Convergence is guaranteed (Theorem 2 in \cite{dempster1977maximum}) although it can be slow.

In this paper the inverse covariance matrix $\mathbf J$ is sparse however the structure of this matrix has no relevance for the derivation of Eq.\ref{e.Mut1}.

For the derivation of $\mathbf J^{t+1}$ we also proceed following the same steps as for the unconstrained full case, with the only attention that the partial derivatives must be only over the non-zero elements with both $i,j$ belonging to a  clique:
\begin{align}%\label{e.MstepE1}
\frac{\partial}{ \partial J_{i,j} } 
&Q(\boldsymbol \mu, {\mathbf J} | \boldsymbol \mu^t, {\mathbf J}^t)    \\
 \!\! &= \!\! \left. \frac12 \sum_{s=1}^q \! \int_0^\infty \!\!\! 
 f( z | \hat {\mathbf x}(s), \boldsymbol \mu^t, {\mathbf J}^t, \nu  )  
  \left(
\!\!-   (\mathbf J^{-1})_{i,j}
\!+\!
z \frac{\nu}{\nu-2} (\hat { x}_i(s) - \mu^t_i)(\hat { x}_{j}(s) - \mu^t_j)
 \right)\!  
dz \! \right|_{\mathbf J = \mathbf J^{t+1}}\!\!\!\!\! \!\!\!\! \!= \! 0 
\end{align}
resulting in the solution
\begin{equation}
 ((\mathbf J^{t+1})^{-1})_{i,j}
\! =  \left(\frac{\nu}{\nu-2} \right) \!\frac1q {\sum_{s=1}^q w_s^t (\hat { x_i}(s) - \mu_i^t)^\top(\hat { x_j}(s) - \mu_j^t)}.
\label{J*MLsolution}
\end{equation}

\end{itemize}

In principle I could perform the EM approach to estimate $\nu$ and, again, sparsity plays no role. 
However, in this paper I prefer to estimate $\nu$ from the tails of the distribution instead of using the EM approach.
Then the computation is reiterated until convergence to a stable set of coefficients.

Let me now proceed with the proofs of theorems  \ref{Th.Smu} and \ref{Th.JMLSt} which are straightforward consequences of the previous derivation. 

\noindent
{\bf Proof of theorem \ref{Th.Smu}}
In order to prove Theorem \ref{Th.Smu}, I must demonstrate that Eqs.\ref{e.StEMmu} and \ref{e.StEMweights} are indeed the maximum likelihood solutions. 
However this is already derived in Eqs.\ref{e.Mut1} and \ref{e.StEMweights1}, providing that the recursion procedure is convergent, but this ie guaranteed by Theorem 2 in \cite{dempster1977maximum}.
$\square$

\noindent
{\bf Proof of theorem \ref{Th.JMLSt}}
Again, in order to prove Theorem \ref{Th.JMLSt}, I must demonstrate that Eqs.\ref{J*S} and \ref{S*} are  the maximum likelihood solutions. 
However this is already derived in Eqs.\ref{J*MLsolution} and \ref{e.StEMweights1}, providing that the recursion procedure is convergent, but this ie guaranteed by Theorem 2 in \cite{dempster1977maximum}.
$\square$

\begin{figure}
\begin{center}
\includegraphics[width=0.7\textwidth]{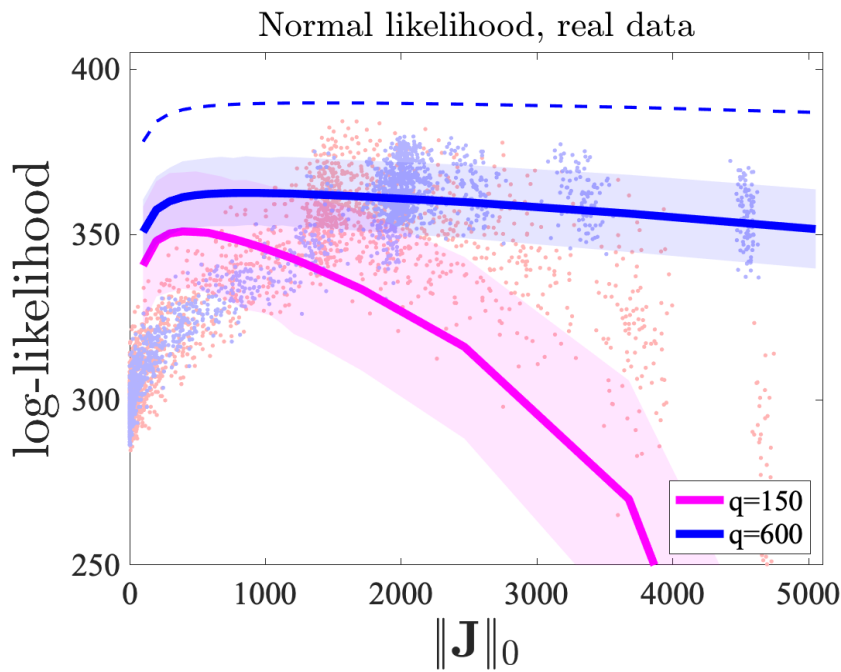}
\caption{
Normal log-likelihood $\ell(\boldsymbol\mu,\mathbf{J} ,\nu)/q$ (Eq.\ref{LLN}) for real financial data (see text) for a set of models with sparse inverse covariance matrix $\mathbf{J}$ constructed by using the MFCF approach with IFN graphs with different levels of sparsity obtained by changing the maximum clique size from 2 to 100.
The IFN have been constructed using the Pearson estimate of the correlation matrix.
The x-axis reports $\left\lVert \mathbf{J}  \right\rVert_0$ which is the number of edges in the IFN graph.
Parameters are estimated on two train sets with different lengths: $q=150$ and $q=600$ respectively (blue and magenta).
Reported results are the log-likelihoods computed on the test set.
The lines are the means and the bands around the lines are the 10\% and 90\% quantiles over 100 random re-sampling.  
The points are instead normal log-likelihood for Glasso models computed using  QUIC using a range of regularization penalty between  $10^{-6}$ to $10^{-3}$.
The slashed blue line is the Student-t likelihood for $q=600$ reported in Fig.\ref{f.TrainValidationResults} which is reported for comparison.
%http://www.cs.utexas.edu/~sustik/QUIC/.
\label{f.RealNormal}}
\end{center}
\end{figure}

\section{Further comparison between models}\label{A.Comparison}

Let me here report some extra results useful for comparison between the models.

I first investigate the real data using the normal modeling instead of the Student-t.
In Fig.\ref{f.RealNormal} I report the log-likelihoods for real data obtained from normal models (Eq.\ref{LLN}) with IFN constructed using the Pearson estimate of the correlation matrix.
The result for the Student-t is reported in this figure with the slashed line for comparison. 
We observe an overall behavior very similar to what reported for the Student-t approach (see Fig.\ref{f.TrainValidationResults}), however the values of the log-likelihoods are significantly lower.
This indicates that real data from financial log-returns are better modeled with Student-t multivariate probability distributions. 
This is not a surprise since the literature abundantly reports the inadequacy of normal modeling for financial returns, yet this result is very clean and has a referential value.

\begin{figure}
\begin{center}
\includegraphics[width=0.7\textwidth]{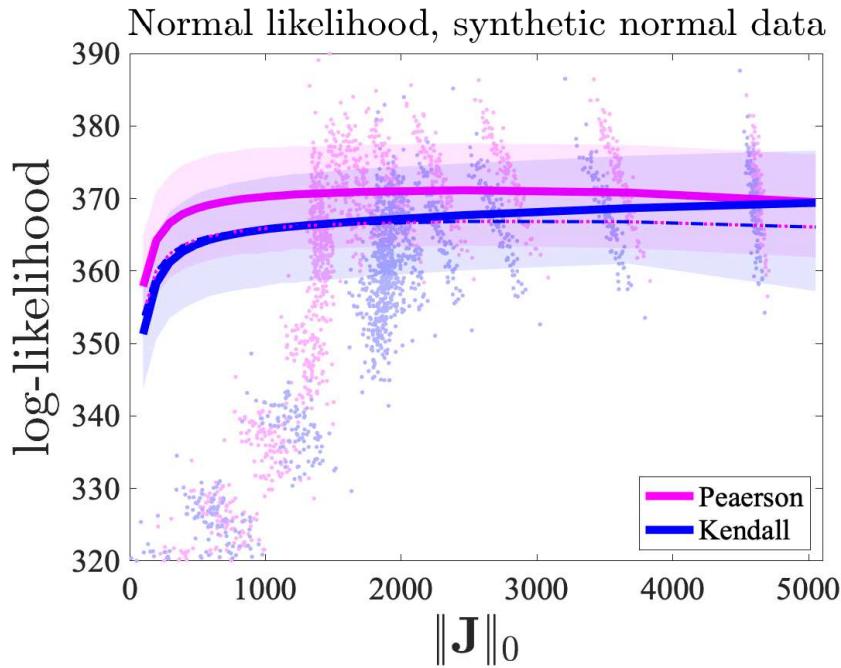}
\caption{
Normal log-likelihood $\ell(\boldsymbol\mu,\mathbf{J} ,\nu)/q$ (Eq.\ref{LLN}) for multivariate normal synthetic data (see text) for a set of models with sparse inverse covariance matrix $\mathbf{J}$ constructed by using the MFCF approach with IFN graphs with different levels of sparsity obtained by changing the maximum clique size from 2 to 100.
The IFN have been constructed using both the Pearson and the Kendall estimates of the correlation matrix (magenta and blue lines respectively).
The x-axis reports $\left\lVert \mathbf{J}  \right\rVert_0$ which is the number of edges in the IFN graph.
Parameters are estimated on train sets of lengths $q=600$.
Reported results are the log-likelihoods computed on the test set.
The lines are the means and the bands around the lines are the 10\% and 90\% quantiles over 100 random re-sampling. 
The points are instead normal log-likelihood for Glasso models computed using  QUIC  using a range of regularization penalty between  $10^{-6}$ to $10^{-3}$.
The slashed blue line and the dotted magenta line are the Student-t models with Kendall and Pearson estimates respectively; they overlap but do not coincide.
%http://www.cs.utexas.edu/~sustik/QUIC/.
\label{f.SynteticNormal}}
\end{center}
\end{figure}

\begin{figure}
\begin{center}
\includegraphics[width=0.7\textwidth]{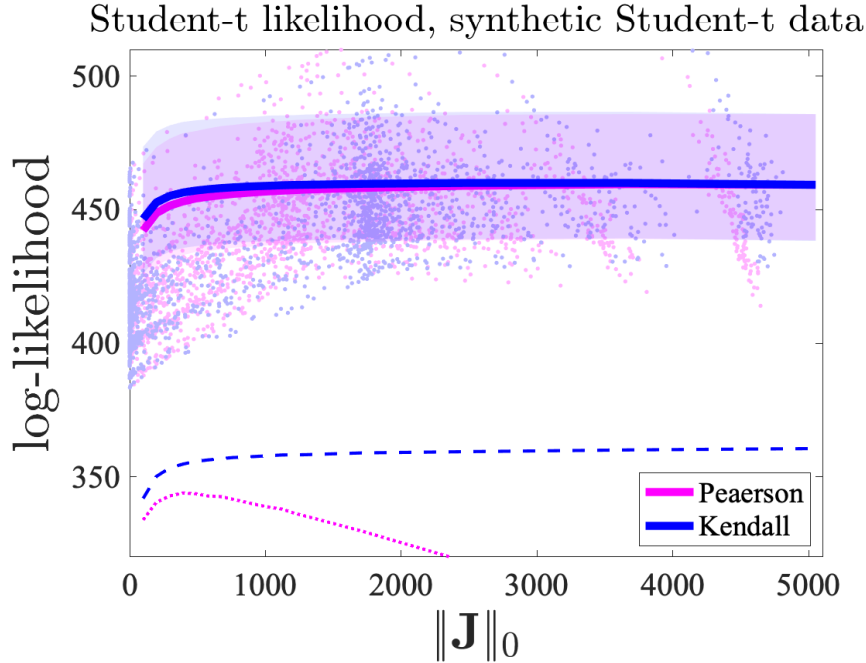}
\caption{
Student-t log-likelihood $\ell(\boldsymbol\mu,\mathbf{J} ,\nu)/q$ (Eq.\ref{LLS}) for multivariate Student-t synthetic data (see text) for a set of models with sparse inverse covariance matrix $\mathbf{J}$ constructed by using the MFCF approach with IFN graphs with different levels of sparsity obtained by changing the maximum clique size from 2 to 100.
The IFN have been constructed using both the Pearson and the Kendall estimates of the correlation matrix (magenta and blue lines respectively).
The x-axis reports $\left\lVert \mathbf{J}  \right\rVert_0$ which is the number of edges in the IFN graph.
Parameters are estimated on train sets of lengths $q=600$.
Reported results are the log-likelihoods computed on the test set.
The lines are the means and the bands around the lines are the 10\% and 90\% quantiles over 100 random re-sampling. 
The points are instead normal log-likelihood for Glasso models computed using  QUIC using a range of regularization penalty between  $10^{-6}$ to $10^{-3}$.
The slashed blue line and the dotted magenta line are the normal models with Kendall and Pearson estimates respectively.
%http://www.cs.utexas.edu/~sustik/QUIC/.
\label{f.SynteticStudentTl}}
\end{center}
\end{figure}

I then repeated the experiments on synthetic data generated from multivariate normal and Student-t distributions.
Results for normally distributed data are reported  in Fig.\ref{f.SynteticNormal}. 
Unsurprisingly, I observe that normal modeling with Pearson estimate of the covariance gives largest likelihoods on normal data. % and that the  is better performing than the Kendall estimate. 
I also observe that the Student-t model with expectation maximization give good results similar to the normal models with Kendal estimate of the covariance. 
For the t model with expectation maximization optimization I obtain very small differences when the Pearson's or Kendall's estimates are used. 
Indeed the slashed and dotted lines appear overlapping, they are however not coinciding and the Pearson's estimate give marginally better results.

Results for Student-t distributed data are reported  in Fig\ref{f.SynteticStudentTl}. 
Here, coherently, I observe that Student-t models largely outperform normal models. 
I also observe that contrary to the normal data case, here the Kendal's estimates of the covariances is advantageous.

%and \ref{f.SynteticStudentTl} for $q=600$ and with estimates of the IFNs performed by using both the Pearson's and Kendal's estimates of the covariances. 
%In the figures I also report with slashed blue line and the dotted magenta line are the Student-t models with Kendall and Pearson estimates respectively
%We observe again a very consistent behaviours with normal models providing better results for normal data and Student-t models giving instead better results for Student-t data. 
%We see that Kendal's estimates of the covariances is advantageous for Student-t models but not for normal models.

In all the cases I have studied Glasso is outperformed by IFN-LoGo sparse models up to a certain level of sparsity and then they become equivalent.

\end{document}